\documentclass[conference]{IEEEtran}
\IEEEoverridecommandlockouts

\usepackage{cite}
\usepackage{amsmath,amssymb,amsfonts}
\usepackage{hyperref}
\usepackage[ruled,noend,linesnumbered]{algorithm2e}
\usepackage{listings}
\usepackage{graphicx}
\usepackage{textcomp}
\usepackage{slashbox}
\usepackage{xcolor}
\usepackage{bm, amsmath}
\def\BibTeX{{\rm B\kern-.05em{\sc i\kern-.025em b}\kern-.08em
    T\kern-.1667em\lower.7ex\hbox{E}\kern-.125emX}}
\begin{document}

\title{Classification with Incoherent Kernel \\ Dictionary Learning
\thanks{This work was supported by a grant of the Romanian Ministry of Education and Research, CCCDI - UEFISCDI, project number PN-III-P2-2.1-PED-2019-3248, within PNCDI III.}}

\author{
\IEEEauthorblockN{Denis C. ILIE-ABLACHIM}
\IEEEauthorblockA{
\textit{Faculty of Automatic Control and Computers},\\ \textit{ University Politehnica of Bucharest}\\
denis.ilie\_ablachim@upb.ro}
\and
\IEEEauthorblockN{Bogdan DUMITRESCU}
\IEEEauthorblockA{
\textit{Faculty of Automatic Control and Computers},\\ \textit{ University Politehnica of Bucharest}\\
bogdan.dumitrescu@upb.ro}}

\maketitle

\begin{abstract}
In this paper we present a new classification method based on Dictionary Learning (DL).
The main contribution consists of a kernel version of incoherent DL, derived from its standard linear counterpart.
We also propose an improvement of the AK-SVD algorithm concerning the representation update.
Our algorithms are tested on several popular databases of classification problems.
\end{abstract}


\section{Introduction}
Dictionary Learning (DL) is a representation learning method used in signal processing and machine learning that aims to find a sparse representation for input data organized as vectors. DL has many applications starting from simple ones like image denoising, inpainting or signal reconstruction and going to coding, clustering or classification. For a given set of samples, $\bm{Y}$, represented by a matrix of $N$ columns (signals) of size $m$, we intend to find a dictionary $\bm{D}$ of size $m \times n$ and a sparse representation matrix $\bm{X}$ of size $n \times N$ such that good sparse representations $\bm{Y} \approx \bm{D} \bm{X}$ are obtained. The representation is based on linear combinations of the columns of the dictionary $\bm{D}$, named atoms. The DL problem can be formulated as follows
\begin{equation}
\begin{array}{ll}
\displaystyle\min_{\bm{D}, \bm{X}} & \|\bm{Y}-\bm{D} \bm{X}\|_{F}^{2} \\
\text { s.t. } & \left\|\bm{x}_{\ell}\right\|_{0} \leq s, \ell=1:N \\
& \left\|\bm{d}_{j}\right\|=1, j=1:n,
\end{array}
\label{DL}
\end{equation}
where $\left\|\cdot\right\|_{0}$ represents the $0$-pseudo-norm and $s$ is the sparsity level.
More precisely, each signal is represented as a linear combination of at most $s$ atoms.

There are several successful DL methods, including K-singular value decomposition (K-SVD) \cite{k-svd} and the Method of Optimal Directions (MOD) \cite{EAH99mod}; improved methods and variations of the DL problem including regularization and coherence reduction are presented in \cite{dlaa}.
All these algorithms are iterative and in most of them an iteration consists of computing the sparse representations $\bm{X}$ with fixed dictionary $\bm{D}$ and then updating the atoms successively, possibly together with the coefficients with which an atom contributes to representations.
Of special interest is the Approximate version of K-SVD (AK-SVD) \cite{RZE08}, which does not seek exact optimality for both an atom and its representation coefficients, but optimizes them successively.
AK-SVD has lower complexity than other algorithms and gives similar end results in most DL problems.

In this paper we present a new perspective on a classification problem via dictionary learning with incoherent atoms. This problem was first introduced in \cite{idl}, where the solution is computed by optimizing the whole dictionary. We introduce a new optimization method in AK-SVD style, in which the dictionary $\bm{D}$ is updated atom by atom. Our contribution is to extend the problem by projecting the signals in a nonlinear space, as linear spaces can hinder classification performance.
To this purpose, we use kernel representations in order to better quantify the similarity between signals.
Another contribution is to introduce a new update rule for the coefficients representations, by taking into consideration only the most recent atoms in all computations; this improvement can lead to the increase of classification accuracy.

The contents of this paper is as follows.
In Section \ref{sec:DLC} we introduce the classification problem and the principle of its solution via DL.
Section \ref{sec:IDL} presents an incoherent DL algorithm suited for classification.
Section \ref{sec:IKDL} contains our main contribution: the kernel version of the incoherent DL algorithm and the new update rule for representations.

Section \ref{sec:exp} is dedicated to experimental results, obtained by running simulations on three publicly available datasets, namely YaleB, AR Face and Caltech 101.

\section{Classification with Dictionary Learning}

\subsection{Standard Dictionary Learning classification}
\label{sec:DLC}

The representation learning approach \eqref{DL} can be also used in classification problems. Considering a set of feature vectors classes $\bm{Y} = \left[ \bm{Y}_{1}, \ldots , \bm{Y}_{c} , \ldots , \bm{Y}_{C} \right]$, where the columns of matrix $\bm{Y}_c \in \mathbb{R}^{m \times N_c}$ are the vectors belonging to class $c$, we intend to learn a specific dictionary, $\bm{D}_{c}$, for each class. For a given test signal $\bm{y} \in \mathbb{R}^{m}$ the classification is achieved by finding the dictionary with the smallest residual of the representation:
\begin{equation}
c = \underset{i=1:C}{\text{argmin}~} \| \bm{y} - \bm{D}_{i} \bm{x}_{i}\|, \ \text{with~} \left\|\bm{x}_{i}\right\|_{0} \leq s.
\label{argmin}
\end{equation}

\subsection{Incoherent Dictionary Learning classification}
\label{sec:IDL}

In order to improve the classification performance, the problem can be extended by adding discriminative power to each dictionary. By this, we intend to maintain a good sparse representation for its own class while achieving a bad representation for the other classes. A solution for this problem was presented in \cite{dlr} where a penalty term was added to the DL problem, transforming it into
\begin{equation}
\min_{\bm{D}_{i}, \bm{X}_{i}}\sum_{i=1}^{C}\left\|\bm{Y}_{i}-\bm{D}_{i} \bm{X}_{i}\right\|_{F}^{2}+\gamma \sum_{i=1}^{C} \sum_{l \neq i} \left\|\bm{D}_{i}^{\top} \bm{D}_{l}\right\|_{F}^{2}.
\label{IDL}
\end{equation}
The second term introduces an incoherence measure between pairs of dictionaries from different classes. By this formulation we intend to project dictionaries into quasi-orthogonal spaces, while retaining most of their representation ability.

The DL problem \eqref{IDL} can be approximately solved by an approach similar to
Approximated K-SVD \cite{k-svd}.
The optimization consists of an iterative process in which the representations $\bm{X}_i$ and
the dictionaries $\bm{D}_i$ are alternately optimized while all other variables are fixed.
The representations are computed with Orthogonal Matching Pursuit (OMP) \cite{omp} as usual in DL, since the penalty term does not depend on $\bm{X}_i$.
The dictionaries are updated sequentially, atom by atom.
Let us assume that we optimize atom $\bm{d}_j$ from dictionary $\bm{D}_i$.
The optimization problem \eqref{IDL} becomes
\begin{equation}
\min _{\bm{d}_{j}} \left\|\bm{F}_{ij}-\bm{d}_{j} \bm{X}_{j, \mathcal{I}_{j}}\right\|_{F}^{2} + 2\gamma \sum_{l \neq i}  \left\|\bm{D}_{l}^{\top} \bm{d}_{j}\right\|_{F}^{2},
\label{opt_dj}
\end{equation}
where $\bm{F}_{ij}=\left[\bm{Y}_{i}-\sum_{\ell \neq j} \bm{d}_{\ell} \bm{x}_{\ell}^{\top}\right]_{\mathcal{I}_{j}}$ is the representation error when all atoms but $\bm{d}_j$ are considered and $\mathcal{I}_{j}$ denotes the indices of the nonzero positions on the $j$th row of $\bm{X}_i$ (those containing the coefficients of $\bm{d}_j$ in the representations).
The solution has been previously presented in \cite{dlr} and is
\begin{equation}
\bm{d}_{j} \leftarrow \bm{F}_{ij} \bm{x}-2 \gamma \bar{\bm{D}} \bar{\bm{D}}^{\top} \bm{d}_{j},
\label{dj_incoh}
\end{equation}
where we have denoted $\bm{x}=\bm{X}_{j, \mathcal{I}_{j}}$ and $\bar{\bm{D}}=\left[\bm{D}_{1}, \ldots , \bm{D}_{i-1} , \bm{D}_{i+1} , \ldots , \bm{D}_{C} \right]$ is the complementary dictionary to the current one.

The atom update operations of the IDL algorithm based on AK-SVD are summarized in Algorithm \ref{alg:IDL} \cite[Alg.4.2]{dlaa}.
Note that the representations are also updated and that the representation error is manipulated efficiently.

\begin{algorithm}[t]
\SetAlgoLined
\KwData{current dictionary $\bm{D} \in \mathbb{R}^{m \times n}$ \\ \hspace{0.85cm} complementary dictionary $\bar{\bm{D}} \in \mathbb{R}^{m \times (n-1)}$ \\ \hspace{0.85cm} representation matrix $\bm{X} \in \mathbb{R}^{n \times N}$}
\KwResult{updated dictionary $\bm{D}$}
Compute error $\bm{E}=\bm{Y}-\bm{D} \bm{X}$ \\
\For{$j=1$ {\bf to} $n$}{
Modify error: $\bm{F}=\bm{E}_{\mathcal{I}_{j}}+\bm{d}_{j} \bm{X}_{j, \mathcal{I}_{j}}$ \\
Update atom: $\bm{d}_{j}=\bm{F} \bm{X}_{j, \mathcal{I}_{j}}^{\top}-2 \gamma \bar{\bm{D}} \bar{\bm{D}}^{\top} \bm{d}_{j}$ \\
Normalize atom: $\bm{d}_{j} \leftarrow \bm{d}_{j} / \left\|\bm{d}_{j}\right\|$ \\ 
Update representation: $\bm{X}_{j, \mathcal{I}_{j}}^{\top}=\bm{F}^{\top} \bm{d}_{j}$ \\
Recompute error: $\bm{E}_{\mathcal{I}_{j}}=\bm{F}-\bm{d}_{j} \bm{X}_{j, \mathcal{I}_{j}}$ \\
}
\caption{Incoherent AK-SVD Dictionary Update}
\label{alg:IDL}
\end{algorithm}

\subsection{Incoherent Kernel Dictionary Learning classification}
\label{sec:IKDL}

In order to evade the linear character of the representation, kernel dictionary learning (KDL) was introduced in \cite{NPNC13,TRS14}. Through this method, the space of signals is extended to a nonlinear feature vector space. We associate with each signal $\bm{y}$ the feature vector $\varphi(\bm{y})$, where $\varphi(\bm{y})$ is a nonlinear function. The dictionary $\bm{D}$ is also extended to a nonlinear space by $\varphi(\bm{Y})\bm{A}$, where $\bm{A}$ contains the coefficients of the dictionary. So, the DL problem \eqref{DL} is transformed into
\begin{equation}
\begin{array}{ll}
\displaystyle\min_{\bm{A}, \bm{X}} & \|\varphi(\bm{Y})-\varphi(\bm{Y})\bm{A} \bm{X}\|_{F}^{2} \\
\text { s.t. } & \left\|\bm{x}_{\ell}\right\|_{0} \leq s, \ell=1:N \\
& \left\|\varphi(\bm{Y})\bm{a}_{j}\right\|=1, j=1:n.
\end{array}
\label{KDL}
\end{equation}
The problem becomes computationally tractable by the use of Mercer kernels, which allows the substitution of the scalar product of feature vectors with the computation of a kernel function $k(\bm{x},\bm{y}) = \varphi(\bm{y})^\top \varphi(\bm{x})$.
Denoting $\bm{K}_{il} = \varphi(\bm{Y}_l)^\top \varphi(\bm{Y}_i)$, the incoherent DL problem \eqref{IDL} is transformed into the Incoherent Kernel Dictionary Learning (IKDL) problem
\begin{equation}
\min_{\bm{A}_{i}, \bm{X}_{i}}\sum_{i=1}^{C}\left\|\varphi(\bm{Y}_{i}) - \varphi(\bm{Y}_i) \bm{A}_{i} \bm{X}_{i}\right\|_{F}^{2}+\gamma \sum_{i=1}^{C} \sum_{l \neq i}\left\|\bm{A}_{i}^{\top} \bm{K}_{li} \bm{A}_{l}\right\|_{F}^{2}.
\label{IKDL}
\end{equation}
Using a similar alternate optimization technique and similar notations, the kernel correspondent of problem \eqref{opt_dj} for optimizing an atom $\bm{a}_j$ is
\begin{align}
\begin{split}
\min_{\bm{a}_{j}}\left\|\varphi(\bm{Y}_{i})\bm{F}_{ij}-\varphi(\bm{Y}_{i})\bm{a}_{j} \bm{X}_{j, \mathcal{I}_{j}}\right\|_{F}^{2} + \\  2\gamma \sum_{l \neq i}  \left\|\bm{A}_{l}^{\top} \bm{K}_{il} \bm{a}_{j}\right\|_{F}^{2}.
\label{ikdl_of}
\end{split}
\end{align}
In order to solve this optimization problem, we compute the partial derivatives with respect to atom $\bm{a}_j$ as follows:
\begin{equation}
\frac{\partial\left\|\varphi(\bm{Y}_{i})\left(\bm{F}_{ij}-\bm{a}_{j} \bm{x}^{\top}\right)\right\|_{F}^{2}}{\partial \bm{a}_{j}}=-2 \bm{K}_{ii} \left(\bm{F}_{ij}-\bm{a}_{j} \bm{x}^{\top}\right) \bm{x}
\label{pd1}
\end{equation}
and
\begin{equation}
\frac{\partial\left\|\bm{A}_{l}^{\top} \bm{K}_{il} \bm{a}_{j}\right\|_{F}^{2}}{\partial \bm{a}_{j}}
= 2 \bm{K}_{il}^{\top} \bm{A}_{l} \bm{A}_{l}^{\top} \bm{K}_{il} \bm{a}_{j}.
\label{pd2}
\end{equation}

By using \eqref{pd1} and \eqref{pd2}, the minimum in \eqref{ikdl_of} is obtained when
\begin{equation}
-\bm{K}_{ii} \left(\bm{F}_{ij} - \bm{a}_{j} x^{\top}\right) \bm{x} +  2\gamma \sum_{l \neq i}  \bm{K}_{il}^{\top} \bm{A}_{l} \bm{A}_{l}^{\top} \bm{K}_{il} \bm{a}_{j} = 0
\end{equation}
and so the solution is
\begin{equation}
\bm{a}_{j} = \left(\bm{K}_{ii}\|\bm{x}\|^{2}+2 \gamma \sum_{l \neq i}  \bm{K}_{il}^{\top} \bm{A}_{l} \bm{A}_{l}^{\top} \bm{K}_{il}\right)^{-1} \bm{K}_{ii} \bm{F}_{ij} \bm{x}.
\end{equation}
The resulting atom is the solution of a $m \times m$ linear system. Given the complexity of the problem, we seek a more convenient approximation.

We note that, given the atom $\bm{a}_j$, the optimal associated representation in \eqref{ikdl_of} is
$\bm{X}_{j, \mathcal{I}_{j}}^{\top}=\bm{F}_{ij}^{\top} \bm{K}_{ii} \bm{a}_{j}$,
like in the kernel K-SVD algorithm (the penalty does not contain the representation).
We insert this optimal representation in \eqref{ikdl_of} and obtain
\begin{equation}
\min_{\bm{a}_{j}} \left\|\varphi(\bm{Y}_{i})\left(\bm{F}_{ij}-\bm{a}_{j} \bm{a}_{j}^{\top} \bm{K}_{ii} \bm{F}_{ij}\right)\right\|_{F}^{2} + 2 \gamma \left\|\hat{\bm{K}_{i}} \bm{a}_{j}\right\|_{F}^{2},
\label{ajx}
\end{equation}
where
\begin{equation}
\hat{\bm{K}_{i}} = \left[\bm{K}_{i1}^{\top} \bm{A}_{1} \ \ldots \ \bm{K}_{i,i-1}^{\top}\bm{A}_{i-1} \ \bm{K}_{i,i+1}^{\top}\bm{A}_{i+1} \  \ldots \ \bm{K}_{iC}^{\top} \bm{A}_{C} \right]^{\top}.
\end{equation}

Expressing the Frobenius norm via its trace form, the new objective from \eqref{ajx} becomes
\begin{align}
\begin{split}
Tr\left[\left(\bm{F}_{ij}-\bm{a}_{j} \bm{a}_{j}^{\top} \bm{K}_{ii} \bm{F}_{ij}\right)^{\top} \bm{K}_{ii} \left(\bm{F}_{ij}-\bm{a}_{j} \bm{a}_{j}^{\top} \bm{K}_{ii} \bm{F}_{ij}\right)\right] + \\ 2 \gamma Tr\left[\bm{a}_{j}^{\top} \hat{\bm{K}_{i}}^{\top} \hat{\bm{K}_{i}} \bm{a}_{j} \right].
\end{split}
\end{align}
After direct transformations and neglecting the terms that do not depend on $\bm{a}_j$, we are left with the minimization of
\begin{equation}
 - \bm{a}_{j}^{\top} \left(\bm{K}_{ii} \bm{F}_{ij} \bm{F}_{ij}^{\top} \bm{K}_{ii} - 2 \gamma \hat{\bm{K}_{i}}^{\top} \hat{\bm{K}_{i}} \right) \bm{a}_{j}.
\label{min_ikdl}
\end{equation}
The solution is the eigenvector corresponding to the maximum eigenvalue of the matrix
\begin{equation}
\bm{H} = \bm{K}_{ii} \bm{F}_{ij} \bm{F}_{ij}^{\top} \bm{K}_{ii} - 2 \gamma \hat{\bm{K}_{i}}^{\top} \hat{\bm{K}_{i}}.
\end{equation}
Since this is again a high complexity operation, we make a single iteration of the power method on the matrix $\bm{H}$. So, given the current atom $\bm{a}_{j}^{(k)}$ (at iteration $k$), the new atom is
\begin{equation}
\bm{a}_{j}^{(k+1)} = \bm{H} \bm{a}_{j}^{(k)} = \bm{K}_{ii} \bm{F}_{ij} \bm{x} - 2 \gamma \hat{\bm{K}_{i}}^{\top} \hat{\bm{K}_{i}} \bm{a}_{j}^{(k)},
\label{dj_kincoh}
\end{equation}
followed by atom normalization.
We have denoted again $\bm{x}=\bm{X}_{j, \mathcal{I}_{j}}$.
The atom update \eqref{dj_kincoh} is the kernel version of \eqref{dj_incoh}.

The atom update operations of the IKDL algorithm are summarized in Algorithm \ref{alg:KIDL} for a single dictionary (hence the index $i$ has disappeared).
We also propose an improvement with respect to the structure of Algorithm \ref{alg:IDL}.
We note that the representation update uses the most recent version of the current atom; however, the error matrix $\bm{F}$ is computed using the previous version of the atom.
By introducing the most recent version of the atom in the error, the representation update becomes
\begin{equation}
\left(\bm{X}_{j, \mathcal{I}_{j}}^{\top}\right)^{(k+1)} =
\bm{F}^{\top} \bm{K} \bm{a}_j = 
\bm{E}^{\top}_{\mathcal{I}_{j}} \bm{K} \bm{a}_{j}
+ \left(\bm{X}^{\top}_{j, \mathcal{I}_{j}}\right)^{(k)}. 
\end{equation}
Due to normalization, we have $\bm{a}_j^\top \bm{K} \bm{a}_j = 1$ and so this product has disappeared from the second term above.
We name Updated-error AK-SVD (UAK-SVD) this version of the algorithm and we will compare it with the usual AK-SVD update.
The difference is only in the representation updates, step 6 of Algorithms \ref{alg:IDL} and \ref{alg:KIDL}.

\begin{algorithm}[t]
\KwData{kernel matrix $\bm{K} \in \mathbb{R}^{N \times N}$ \\ \hspace{0.85cm} current dictionary $\bm{A} \in \mathbb{R}^{N \times n}$ \\ \hspace{0.85cm} complementary dictionary $\hat{\bm{K}} \in \mathbb{R}^{(N-1) \times N}$ \\ \hspace{0.85cm} representation matrix $\bm{X} \in \mathbb{R}^{n \times N}$}
\KwResult{updated dictionary $\bm{D}$}
Compute error $\bm{E}=\bm{I}-\bm{A} \bm{X}$ \\
\For{$j=1$ {\bf to} $n$}{
Modify error: $\bm{F}=\bm{E}_{\mathcal{I}_{j}} + \bm{a}_{j} \bm{X}_{j, \mathcal{I}_{j}}$  \\
Update atom: $\bm{a}_{j}= \bm{K} \bm{F} \bm{X}_{j, \mathcal{I}_{j}} - 2 \gamma \hat{\bm{K}}^{\top} \hat{\bm{K}} \bm{a}_{j}$ \\
Normalize atom: $\bm{a}_{j} \leftarrow\left(\bm{a}_{j}^{\top} \bm{K} \bm{a}_{j}\right)^{\frac{1}{2}}$ \\ 
Update representation: $\bm{X}_{j, \mathcal{I}_{j}}^{\top} \leftarrow \bm{E}^{\top}_{\mathcal{I}_{j}} \bm{K} \bm{a}_{j} + \bm{X}_{j, \mathcal{I}_{j}}^{\top}$ \\
Recompute error: $\bm{E}_{\mathcal{I}_{j}}=\bm{F}-\bm{a}_{j} \bm{X}_{j, \mathcal{I}_{j}}$ \\
}
\caption{Incoherent Kernel UAK-SVD Dictionary Update}
\label{alg:KIDL}
\end{algorithm}

For the classification scheme we need only the reconstruction errors from equation \eqref{argmin}. For the kernel version, the classification of a signal $\bm{y}$ results from
\begin{equation}
c = \underset{i=1:C}{\text{argmin}~} \| \varphi(\bm{y}) - \varphi(\bm{Y}_{i}) \bm{A}_{i} \bm{x}_{i}\|, \ \text{with~} \left\|\bm{x}_{i}\right\|_{0} \le s,
\end{equation}
which leads to
\begin{align}
\begin{split}
c = \underset{i=1:C}{\text{argmin}~} k(\bm{y},\bm{y}) + \bm{x}_i^{\top} \bm{A}_{i}^{\top} \bm{K}_{i} \bm{A}_{i} \bm{x}_i - 2 k(\bm{y}, \bm{Y}_{i}) \bm{A}_{i} \bm{x}, \\ \text{with~} \left\|\bm{x}_{i}\right\|_{0} \le s.
\end{split}
\end{align}
Here, as well as in the IKDL algorithm, the representations are computed with Kernel OMP \cite{NPNC13}.

\section{Experiments}
\label{sec:exp}

In this section we present the main results obtained with the Incoherent Kernel Dictionary Learning algorithm. The datasets used in the simulations are YaleB \cite{yaleb}, AR Face \cite{arface} and Caltech 101 \cite{caltech101}.

For the evaluation step, each dataset is independently used and was provided in \cite{jiang2013label}. We measure performance through classification accuracy, training time and testing time.

All the algorithms were developed in Matlab $2018a$, on a laptop with $3.5$GHz Intel CPU and $16$ GB RAM memory. The execution time and accuracy are reported as the average over the 3 best results. For the methods that require the use of a kernel function, we used two types of kernels: radial basis function kernel ($k(\bm{x},\bm{y}) = \exp{\frac{-||\bm{x}-\bm{y}||_2^2}{2\sigma^2}}$) and polynomial kernel ($k(\bm{x},\bm{y})=(\bm{x}^{\top}\bm{y}+\alpha)^\beta$). For the kernel methods, we have tried different parameter values in our simulations. We have chosen the final form based on the best results from these simulations. The code for the proposed algorithms is available at \href{https://github.com/denisilie94/Incoherent-Kernel-Dictionary-Learning}{https://github.com/denisilie94/Incoherent-Kernel-Dictionary-Learning}.

\textbf{YaleB Database} is organized into two sub-datasets, according to the extended and cropped images. The dataset is composed of $16128$ images of $38$ human subjects under $9$ poses and $64$ illumination conditions. During the simulation step only the extended dataset was used, including $2414$ face images of $38$ persons. For the training and testing step the images per subject were split in half. The dimension of the feature vectors is $504$.

\textbf{AR Face Database} is a face dataset containing more than $4000$ color images corresponding to $126$ different people ($70$ men and $56$ women). The images were taken having a frontal view with different facial expressions, illumination conditions and occlusions. For the experimental phase a set of $2600$ images of $50$ females and $50$ male subjects are extracted. For each subject, $20$ images were used for training and $6$ for testing.

Beside the face recognition tasks, an object recognition task was attempted in the simulations. For this we used \textbf{Caltech 101 Database}. The dataset includes 9,144 images from 102 classes (101 common object classes and a background class). The number of samples in each category varies from 31 to 800. In the experiments, 30 samples per category were used for training, while the rest are used for testing.

During the simulations we performed tests with dictionaries of different sizes ($40$, $60$, $80$ and $100$ atoms) having a sparsity constraint equal to $10\%$, $20\%$, $50\%$ and $80\%$ of the number of atoms. Taking into account the training time and the resulted classification accuracy, we chose to use only dictionaries with 40 atoms and a sparsity constraint of $20$. Increasing sparsity can improve the results, but this will also affect the training time. All tests were performed on 10 DL iterations. For a larger number of iterations the improvement in accuracy is insignificant. We set the hyperparameters of the optimization problem following a grid search: $\gamma \in [0.01, 0.1, 0.5, 1, 2, 4, 6]$, $\sigma \in [0.5, 1, 2, 4, 5, 6, 8, 10]$, $\alpha \in [0.5, 1, 2, 4]$ and $\beta \in [2, 3]$. In the case of all datasets, for the IDL problem we used $\gamma=4$, while for the IKDL problem $\gamma$ was set to $0.1$. Regarding the kernel functions, we used the following parameters: $\sigma=4$, $\alpha=2$ and $\beta=2$ for YaleB dataset; $\sigma=8$, $\alpha=4$ and $\beta=2$ for AR Face dataset; and $\sigma=5$, $\alpha=4$ and $\beta=2$ for Caltech 101 dataset.

The main results are summarized in Tables \ref{tab:aksvd}, \ref{tab:uaksvd} for classification with plain incoherent DL; Tables \ref{tab:aksvd_rbf}, \ref{tab:uaksvd_rbf} contain results with IKDL and the RBF kernel; Tables \ref{tab:aksvd_poly} and \ref{tab:uaksvd_poly} contain results with IKDL and polynomial kernel. As we can see, the results vary depending on the chosen algorithm. The UAK-SVD method usually improves the classification accuracy, although sometimes only slightly. Regarding the kernel extension, the introduced nonlinearity does not always insure an improvement, as we can see for YaleB dataset, but there is a strong improvement regarding the execution time. In the case of YaleB dataset, the training time decreased by $10$ times, while for the AR Face dataset the training is done $25$ times faster. The best improvement is visible for the Caltech 101 dataset, where training time has been reduced $200$ times. 
The execution time is reduced due to the small size of the dictionaries in the kernel version. This property is valid only for cases where the signal size is much larger than the number of signals per class; for example, in the YaleB case, the dictionary of a class has size $504 \times 40$ in the IDL approach, but size only $32 \times 40$ in IKDL; it is thus remarkable that the accuracy loss is so small when kernels are used. This property is also valid for the other datasets, where we have signals of size 540 for AR Face dataset and 3000 for Caltech 101 dataset.

\begin{figure}[!ht]
\centerline{\includegraphics[width=8cm]{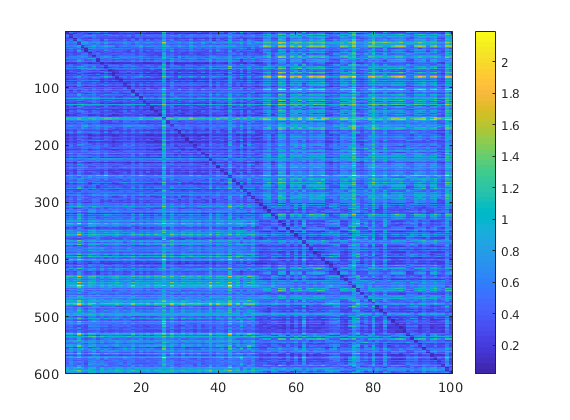}}
\caption{$\left\|\varphi(\bm{y})-\varphi(\bm{Y}_{i})\bm{A}_{i}\bm{x}\right\|_{F}^{2}$ (YaleB - UAK-SVD IKDL)}
\label{fig:yaleb}
\end{figure}

\begin{figure}[!ht]
\centerline{\includegraphics[width=8cm]{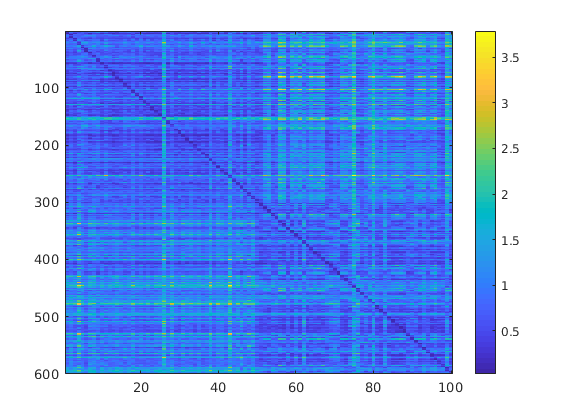}}
\caption{$\left\|\varphi(\bm{y})-\varphi(\bm{Y}_{i})\bm{A}_{i}\bm{x}\right\|_{F}^{2}$ (AR Face - UAK-SVD IKDL)}
\label{fig:arface}
\end{figure}

\begin{figure}[!ht]
\centerline{\includegraphics[width=8cm]{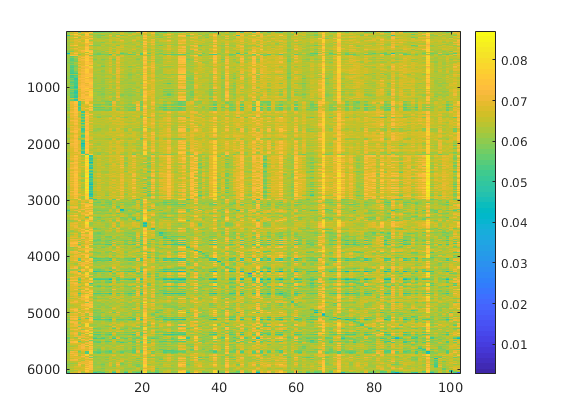}}
\caption{$\left\|\varphi(\bm{y})-\varphi(\bm{Y}_{i})\bm{A}_{i}\bm{x}\right\|_{F}^{2}$ (Caltech101 - UAK-SVD IKDL)}
\label{fig:caltech101}
\end{figure}

\begin{figure}[!ht]
\centerline{\includegraphics[width=8cm]{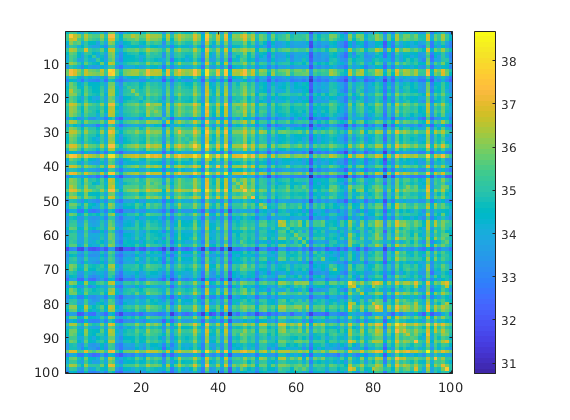}}
\caption{$\left\|\bm{A}_{l}^{\top} \bm{K}_{il} \bm{A}_{i}\right\|_{F}^{2}$ (YaleB - UAK-SVD IKDL)}
\label{fig:yaleb_discr}
\end{figure}

\begin{figure}[!ht]
\centerline{\includegraphics[width=8cm]{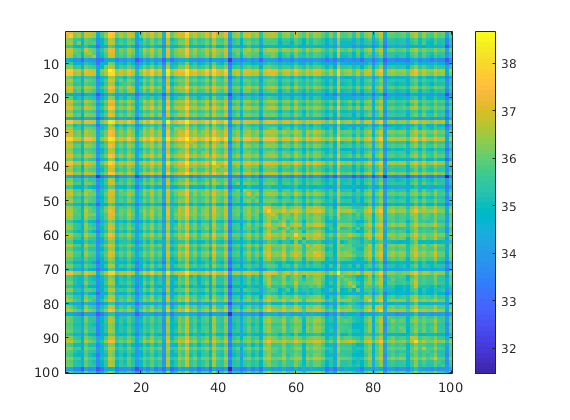}}
\caption{$\left\|\bm{A}_{l}^{\top} \bm{K}_{il} \bm{A}_{i}\right\|_{F}^{2}$ (AR Face - UAK-SVD IKDL)}
\label{fig:arface_discr}
\end{figure}

\begin{figure}[!ht]
\centerline{\includegraphics[width=8cm]{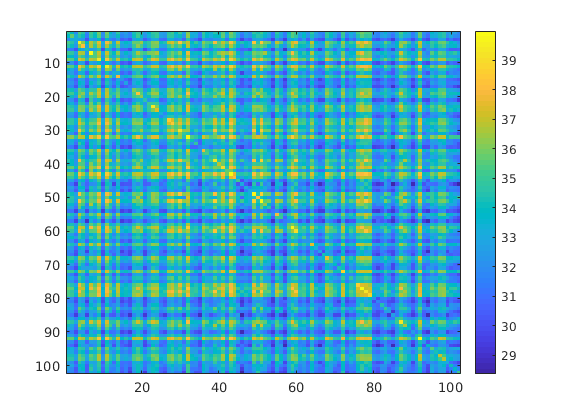}}
\caption{$\left\|\bm{A}_{l}^{\top} \bm{K}_{il} \bm{A}_{i}\right\|_{F}^{2}$ (Caltech101 - UAK-SVD IKDL)}
\label{fig:caltech101_discr}
\end{figure}

In order to better understand the classification problem we compute the reconstruction error (figures \ref{fig:yaleb}, \ref{fig:arface} and \ref{fig:caltech101}) and the discriminative term (figures \ref{fig:yaleb_discr}, \ref{fig:arface_discr} and \ref{fig:caltech101_discr}). Based on the exploitation of the two terms we can easily see that the reconstruction error achieves good representation for YaleB and AR Face datasets, while the discriminative term does not produce the quasi-orthogonality of the dictionaries. For these problems, the classification obtains good results by taking $\gamma$ small enough so that the discriminative term does not have an important weight in the objective function. On the other side, the Caltech 101 dataset does not achieve a separable error reconstruction, but the discriminative term is stronger and thus classification can be performed.

\vspace{-10pt}
\begin{table}[!ht]
\caption{AK-SVD Incoherent Dictionary Learning}
\begin{tabular}{|l||*{4}{c|}}\hline
\backslashbox{Dataset}{Perf.} & \makebox[5.5em]{Train. time} 
& \makebox[5.5em]{Test. time} & \makebox[5.5em]{Accuracy}\\\hline\hline
YaleB & 82.9 [sec] & 20.6 [sec] & 94.00\% \\\hline
AR Face & 558 [sec] & 27 [sec] & 93.22\% \\\hline
Caltech101 & 14332 [sec] & 329 [sec] & 67.30\% \\
\hline
\end{tabular}
\label{tab:aksvd}
\end{table}

\vspace{-18pt}
\begin{table}[!ht]
\caption{UAK-SVD Incoherent Dictionary Learning}
\begin{tabular}{|l||*{4}{c|}}\hline
\backslashbox{Dataset}{Perf.} & \makebox[5.5em]{Train. time} 
& \makebox[5.5em]{Test. time} & \makebox[5.5em]{Accuracy}\\\hline\hline
YaleB & 87.5 [sec] & 20.8 [sec] & 94.11\% \\\hline
AR Face & 560 [sec] & 26.3 [sec] & 93.33\% \\\hline
Caltech101 & 14367 [sec] & 329 [sec] & 66.98\% \\
\hline
\end{tabular}
\label{tab:uaksvd}
\end{table}

\vspace{-18pt}
\begin{table}[!ht]
\caption{AK-SVD Incoherent Kernel Dictionary Learning \\ (RBF kernel)}
\begin{tabular}{|l||*{4}{c|}}\hline
\backslashbox{Dataset}{Perf.} & \makebox[5.5em]{Train. time} 
& \makebox[5.5em]{Test. time} & \makebox[5.5em]{Accuracy}\\\hline\hline
YaleB & 6.8 [sec] & 23.8 [sec] & 93.88\% \\\hline
AR Face & 19.5 [sec] & 25.3 [sec] & 93.42\% \\\hline
Caltech101 & 62.4 [sec] & 403 [sec] & 70.67\% \\
\hline
\end{tabular}
\label{tab:aksvd_rbf}
\end{table}

\vspace{-18pt}
\begin{table}[!ht]
\caption{UAK-SVD Incoherent Kernel Dictionary Learning \\(RBF kernel)}
\begin{tabular}{|l||*{4}{c|}}\hline
\backslashbox{Dataset}{Perf.} & \makebox[5.5em]{Train. time} 
& \makebox[5.5em]{Test. time} & \makebox[5.5em]{Accuracy}\\\hline\hline
YaleB & 7.2 [sec] & 24.7 [sec] & 94.05\% \\\hline
AR Face & 19.2 [sec] & 25.3 [sec] & 93.50\% \\\hline
Caltech101 & 62.7 [sec] & 402 [sec] & 70.12\% \\
\hline
\end{tabular}
\label{tab:uaksvd_rbf}
\end{table}

\vspace{-18pt}
\begin{table}[!ht]
\caption{AK-SVD Incoherent Kernel Dictionary Learning \\(Polynomial kernel)}
\begin{tabular}{|l||*{4}{c|}}\hline
\backslashbox{Dataset}{Perf.} & \makebox[5.5em]{Train. time} 
& \makebox[5.5em]{Test. time} & \makebox[5.5em]{Accuracy}\\\hline\hline
YaleB & 9.0 [sec]  & 26.2 [sec]  & 94.00\% \\\hline
AR Face & 24.5 [sec] & 30.5 [sec] & 94.83 \% \\\hline
Caltech101 & 73.3 [sec] & 430 [sec] & 70.83\%  \\
\hline
\end{tabular}
\label{tab:aksvd_poly}
\end{table}

\vspace{-18pt}
\begin{table}[!ht]
\caption{UAK-SVD Incoherent Kernel Dictionary Learning \\(Polynomial kernel)}
\begin{tabular}{|l||*{4}{c|}}\hline
\backslashbox{Dataset}{Perf.} & \makebox[5.5em]{Train. time} 
& \makebox[5.5em]{Test. time} & \makebox[5.5em]{Accuracy}\\\hline\hline
YaleB & 9.0 [sec]  & 27.1 [sec]  & 94.07\% \\\hline
AR Face & 24.5 [sec] & 30.3 [sec] & 94.83\% \\\hline
Caltech101 & 73.6 [sec] & 428 [sec] & 71.46\% \\
\hline
\end{tabular}
\label{tab:uaksvd_poly}
\end{table}

\newpage

\section{Conclusions}
In this paper we have extended the family of dictionary learning algorithms for classification problems.
We have presented a modified version of AK-SVD in which the most recent version of an atom is used in all respects in the representation update.
We have proposed a kernel version of incoherent AK-SVD that can improve classification performance by increasing the separation of dictionaries dedicated to different signal classes.
The experimental results confirm the good behavior of our algorithms, especially in terms of complexity.

\bibliographystyle{unsrt}
\bibliography{bib}

\end{document}